\documentclass[11pt]{article}
\usepackage[final]{acl}

\usepackage{times}
\usepackage{latexsym}
\usepackage{booktabs}
\usepackage{tabularx}
\usepackage{comment}
\usepackage{booktabs}
\usepackage[T1]{fontenc}
\usepackage[utf8]{inputenc}

\usepackage{microtype}
\usepackage{inconsolata}

\usepackage{graphicx}

\usepackage{amsmath,amssymb,amsfonts}
\usepackage{booktabs}
\usepackage{array}
\usepackage{multirow}

\title{IntelliCode: A Multi-Agent LLM Tutoring System with Centralized Learner Modeling}

\author{
\textbf{Jones David\textsuperscript{1}},
\textbf{Shreya Ghosh\textsuperscript{2}}
\\
\\
\textsuperscript{1}School of Computer Science and Engineering, VIT-AP University, India \\
\textsuperscript{2}School of Electrical and Computer Sciences, Indian Institute of Technology Bhubaneswar, India
\\
\small{
\textbf{Correspondence:}
\href{mailto:jones.22bce8135@vitapstudent.ac.in}{jones.22bce8135@vitapstudent.ac.in},
\href{mailto:shreya@iitbbs.ac.in}{shreya@iitbbs.ac.in}
}
}

\begin{document}

\maketitle

\begin{abstract}
LLM-based tutors are typically single-turn assistants that lack persistent representations of learner knowledge, making it difficult to provide principled, transparent, and long-term pedagogical support. We introduce \textbf{IntelliCode}, a multi-agent LLM tutoring system built around a centralized, versioned learner state that integrates mastery estimates, misconceptions, review schedules, and engagement signals. A StateGraph Orchestrator coordinates six specialized agents: skill assessment, learner profiling, graduated hinting, curriculum selection, spaced repetition, and engagement monitoring, each operating as a pure transformation over the shared state under a single-writer policy. This architecture enables auditable mastery updates, proficiency-aware hints, dependency-aware curriculum adaptation, and safety-aligned prompting. Our demo showcases an end-to-end tutoring workflow: a learner attempts a DSA problem, receives a conceptual hint when stuck, submits a corrected solution, and immediately sees mastery updates and a personalized review interval. We report validation results with simulated learners, showing stable state updates, improved task success with graduated hints, and diverse curriculum coverage. 
\footnote{\textbf{Live System}: \url{https://intellicode.redomic.in}}
\footnote{\textbf{Video Demo}: \url{https://youtu.be/oO8bZfeleOU}}
\end{abstract}

\section{Introduction}

Large Language Models (LLMs) have rapidly expanded the possibilities of automated tutoring, yet most existing systems remain fundamentally reactive: each query is treated in isolation, with little continuity or awareness of a learner’s evolving knowledge~\cite{wu2025embracingimperfectionsimulatingstudents}. Human tutors, in contrast, maintain rich, persistent models of student understanding that support targeted feedback, curriculum scaffolding, and long-term learning trajectories~\cite{VanLEHN01102011}. This discrepancy limits the pedagogical reliability of current LLM tutors, which often provide inconsistent hints, overlook dependencies between concepts, and fail to account for systematic misconceptions.
\begin{figure}[t]
\centering
\includegraphics[width=0.7\linewidth]{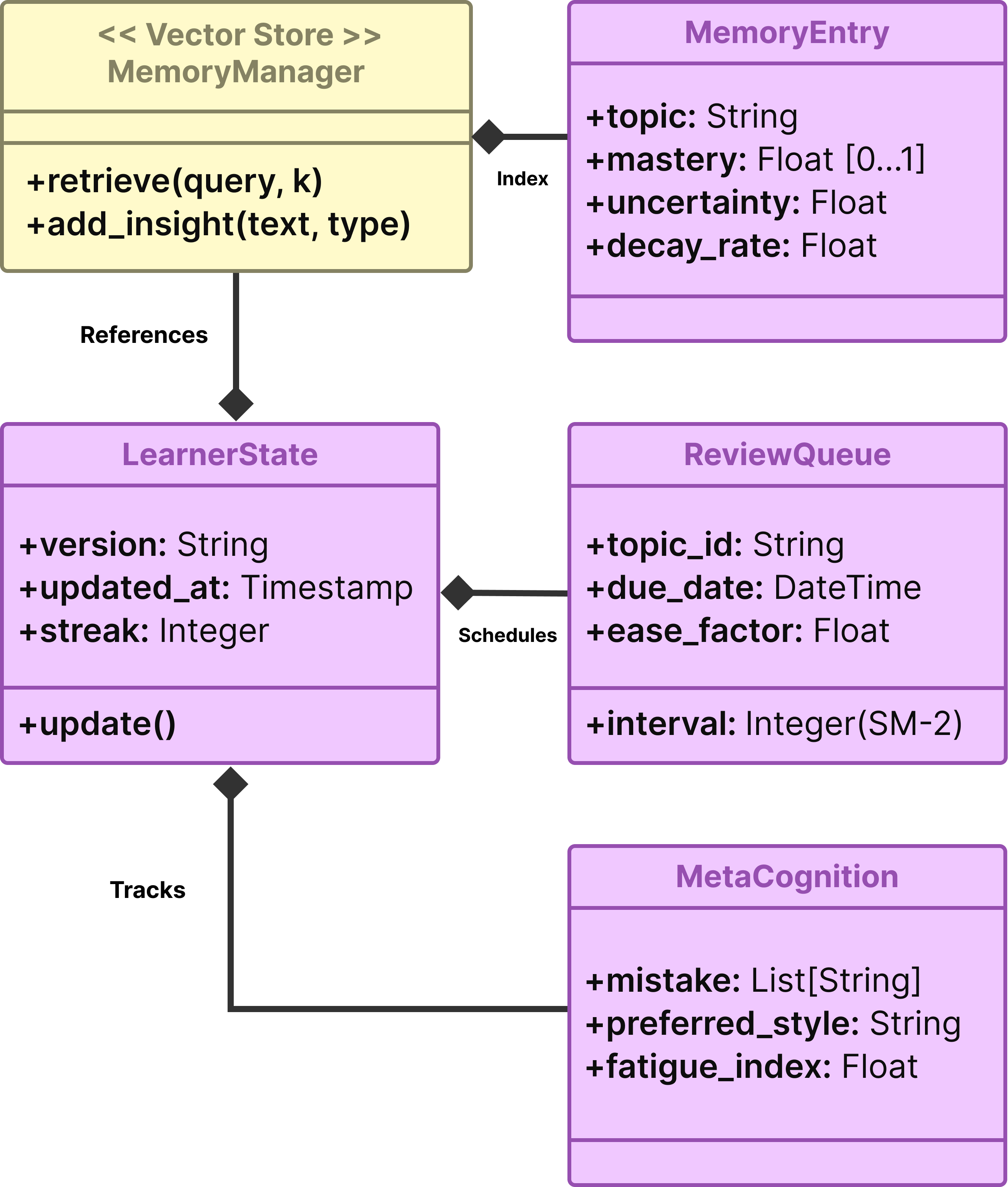}
\caption{Unified Learner State Schema.}
\label{fig:schema}
\end{figure}
Recent frameworks such as GenMentor~\cite{wang2025llmpoweredmultiagentframeworkgoaloriented} and SocraticLM~\cite{liu2024socraticlm} demonstrate the promise of multi-agent orchestration and structured dialogue in stabilizing tutoring behavior. However, these systems typically rely on ephemeral or implicit memory, lacking an explicit, auditable learner model shared across agents. At the same time, decades of work on learner modeling from Bayesian Knowledge Tracing (BKT)~\cite{corbett1995knowledge} and Deep Knowledge Tracing (DKT)~\cite{piech2015deep} to memory networks~\cite{zhang2017dynamic} and PFA~\cite{pavlik2009performance}, underscore the importance of accurate mastery estimation for personalization. Yet, few LLM-based tutors integrate such formal models with generative reasoning in a persistent, updateable state.

\textbf{Our goal is to bridge this gap.} We introduce \textbf{IntelliCode}, a multi-agent LLM tutoring system built around a centralized, versioned learner state that serves as the single source of truth for all pedagogical decisions. Unlike prior systems, IntelliCode enforces a single-writer policy through a \textit{StateGraph Orchestrator}, ensuring that every mastery update, hint intervention, or curriculum choice results from a coherent, formally validated transformation of the learner model. This design mitigates drift, prevents conflicting agent outputs, and enables transparent, multi-turn personalization.

IntelliCode integrates well-established instructional principles: mastery estimation, graduated hinting, dependency-aware curriculum planning, and spaced repetition; with modern LLM capabilities. The learner state encodes mastery vectors with uncertainty, misconceptions, review schedules, and behavioral signals. Six specialized agents (Skill Assessment, Learner Profiler, Pedagogical Feedback, Content Curator, Progress Synthesizer, and Engagement Orchestrator) each operate as pure functions over this shared state. For example, if a learner repeatedly omits a base case in recursion problems, the Profiler records a misconception, the Feedback agent adapts its hint level, and the Curator adjusts upcoming tasks accordingly.

\paragraph{Contributions.}
This work makes the following contributions:
(1) a centralized, versioned learner state and single-writer orchestration mechanism enabling consistent, auditable multi-turn tutoring;
(2) a suite of pedagogical agents that implement mastery estimation, graduated hinting, curriculum planning, and spaced repetition as pure transformations over the shared state; and
(3) a fully functional, end-to-end LLM tutoring system demonstrating stable interaction, interpretable decision-making, and robust content coverage through simulated learner studies.

Figure~\ref{fig:schema} highlights the unified learner state that drives our adaptation policies. By grounding agent behavior in this structured representation while leveraging LLM reasoning for high-variance tasks such as hinting and code analysis, IntelliCode offers a transparent and pedagogically consistent alternative to memory-less conversational tutoring.

\section{System Architecture}\label{sec:architecture}

We frame adaptive personalized education as a Partially Observable Markov Decision Process (POMDP). At each timestep, the learner state $\mathbf{S}_t$ maintains mastery vectors, SM-2--based review schedules, engagement metrics, and metacognitive memory. Observations $\mathcal{O}_t$ reflect noisy behavioral signals such as submissions, errors, and hint requests, while actions $\mathcal{A}_t$ correspond to pedagogically meaningful interventions including content recommendation, graduated hinting, and schedule adjustments. The reward function $R_t$ trades off mastery gains with penalties for excessive hint usage and inefficient solve times (details in Appendix~\ref{app:formulation}). This formulation enables principled adaptation while supporting the modular multi-agent design showcased in the demo.
\begin{table*}[h]
\scriptsize
\centering
\caption{Roles and responsibilities of the six pedagogical agents.}
\label{tab:agent_roles}
\begin{tabular}{p{0.52\columnwidth} p{1.5\columnwidth}}
\toprule
\textbf{Agent} & \textbf{Responsibility} \\
\midrule
Pedagogical Feedback & Provides proficiency-aware, five-level graduated hinting without solution disclosure. \\ \midrule
Content Curator & Selects personalized problems based on mastery, dependencies, and the 40/50/10 curriculum policy. \\ \midrule
Engagement Orchestrator & Monitors motivation, pacing, and disengagement signals to issue supportive nudges. \\ \midrule
Skill Assessment & Performs hybrid evaluation using test-case execution and semantic code review. \\ \midrule
Learner Profiler & Estimates mastery deltas, identifies misconceptions, and infers behavioral trends. \\ \midrule
Progress Synthesizer & Schedules reviews using an enhanced SM-2 mechanism with context-aware adjustments. \\
\bottomrule
\end{tabular}
\end{table*}
\subsection{Orchestrator Overview}

At the core of IntelliCode is the \textbf{StateGraph Orchestrator}, the only component permitted to write to the persistent learner record. It maintains a synchronized in-memory copy of the learner state and coordinates all interactions among the six pedagogical agents. When an event occurs, the orchestrator routes it to the appropriate agents, aggregates their outputs, validates the proposed state changes, and then commits them as an atomic update. Similar to the coordination strategy in GenMentor~\cite{wang2025llmpoweredmultiagentframeworkgoaloriented}, this mechanism prevents conflicting writes, enforces safety and schema constraints, and ensures that the system behaves predictably over multi-turn, long-term tutoring sessions. The orchestrator thus provides the reliability and auditability necessary for principled learner modeling.

\subsection{Trigger Types and Routing}

The orchestrator reacts to pedagogically meaningful events and dispatches them to the relevant agents. These triggers operationalize the full workflow demonstrated in the system:

\begin{itemize}
    \item \textbf{on\_submission}: A code or answer submission triggers Skill Assessment, followed by the Learner Profiler and Pedagogical Feedback.
    \item \textbf{on\_hint\_request}: A learner request for help triggers the Pedagogical Feedback agent, informed by current proficiency and hint history.
    \item \textbf{on\_session\_check}: A daily check-in triggers the Content Curator and Engagement Orchestrator.
    \item \textbf{on\_daily\_generation}: The system generates the day's personalized problem set via the Content Curator.
    \item \textbf{on\_review\_due}: When an SM-2 review is due, the Progress Synthesizer and Content Curator are invoked.
\end{itemize}

These triggers allow IntelliCode to respond adaptively to the learner’s evolving behavior, maintaining pedagogical continuity across sessions.

\subsection{Overview of System Agents}

\begin{figure*}[t]
\centering
\includegraphics[width=0.7\textwidth]{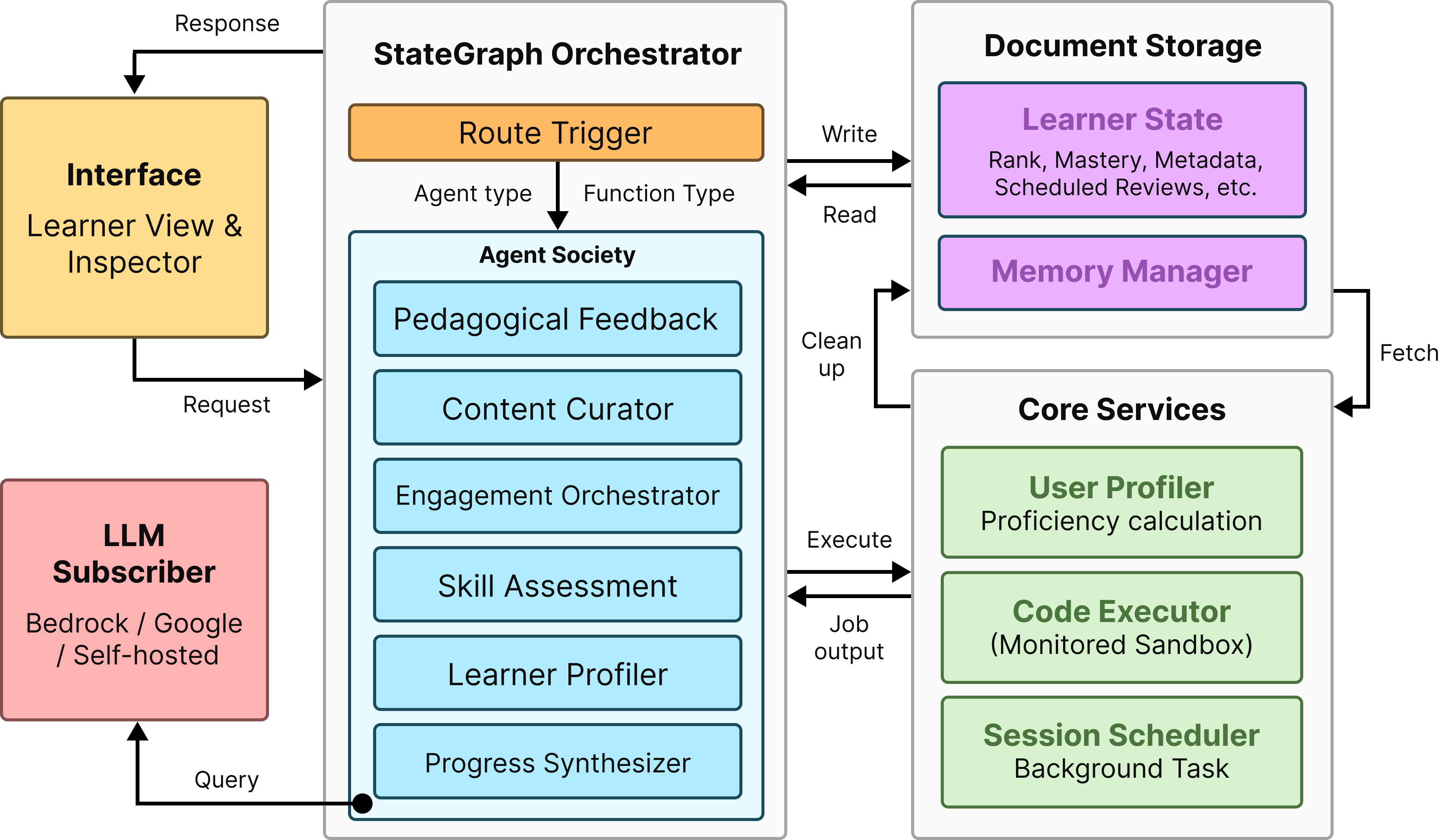}
\caption{System Architecture: The StateGraph Orchestrator manages the flow between six specialized agents and the persistent learner state. Arrows indicate data flow and trigger events.}
\label{fig:architecture}
\end{figure*}

Each agent operates as a pure transformation over the shared learner state, producing structured outputs that the orchestrator validates and integrates. Together, the six agents support assessment, personalization, pacing, hinting, and review scheduling. Table~\ref{tab:agent_roles} summarizes their responsibilities, and Figure~\ref{fig:architecture} shows how the orchestrator mediates their communication. This design ensures that all instructional decisions are grounded in a consistent, auditable learner state and remain traceable throughout the tutoring trajectory.

The overall data flow is illustrated in Figure~\ref{fig:architecture}. The StateGraph Orchestrator mediates all communication between agents and the persistent learner state, ensuring that instructional decisions remain traceable and consistent across sessions.

\section{Learner State and Agent Adaptation}
\label{sec:learner_state}

IntelliCode maintains a centralized, versioned learner state that governs all pedagogical decisions. The state is initialized from historical activity and updated using a Bayesian Knowledge Tracing (BKT)--inspired mechanism that incorporates difficulty, recency, hint usage, and solve-time effects. Spaced repetition is managed by an enhanced SM-2 scheduler that computes personalized review intervals based on recall quality and interaction history. This shared representation enables consistent, long-term adaptation rather than isolated single-turn responses.

To illustrate the update process, consider a learner who solves a recursion problem correctly but requests several hints and exceeds the expected solve time. The BKT-inspired update assigns a modest mastery gain, attenuated by the heavier reliance on hints, while the Progress Synthesizer schedules an earlier review to reinforce retention. The Content Curator then interprets recursion as lying in the learner’s “growth” region and adjusts future problem selection accordingly.

\subsection{Agent Behaviors and Pedagogical Logic}

All six agents operate as pure transformations over the learner state, producing structured outputs validated by the orchestrator before being committed as atomic updates. While the architecture supports fully generative agents, deterministic logic is used for the Learner Profiler and Content Curator for reproducibility, whereas higher-variance components (e.g., hinting, code analysis) leverage LLMs.

\paragraph{Learner Profiler}
The Profiler acts as the diagnostic backbone of the system, identifying mastery deltas, misconceptions, and behavioral trends such as fatigue or decreasing velocity. It consumes correctness, topic tags, error patterns, time-on-task, hint usage, and the current mastery map. For example, if a learner repeatedly omits base cases in recursion problems, the Profiler records a misconception related to termination conditions, which later guides both hinting and content selection.

\paragraph{Skill Assessment}
This agent performs hybrid evaluation by executing test cases and conducting semantic code review. When tests fail, errors alone are surfaced; when they pass, the agent provides improvement suggestions across categories such as time complexity, space usage, readability, and edge-case coverage. For instance, after a successful merge sort implementation, the agent may recommend reducing auxiliary memory to improve space efficiency.

\paragraph{Pedagogical Feedback}
Informed by Socratic Playground~\cite{zhang2024splsocraticplaygroundlearning}, the Pedagogical Feedback agent employs a five-level graduated hinting protocol:

\begin{itemize}
    \item \textbf{Metacognitive:} prompt the learner to reflect (“What did you try, and what happened?”).
    \item \textbf{Conceptual:} surface the key idea (“This problem relies on identifying a recurrence relation.”).
    \item \textbf{Strategic:} suggest an approach (“Consider breaking the input and solving the halves recursively.”).
    \item \textbf{Structural:} highlight missing logic (“Your solution lacks a base case for empty input.”).
    \item \textbf{Targeted:} point to a region of interest (“Inspect the condition near line 14; termination may not be guaranteed.”).
\end{itemize}
\begin{figure*}[t]
\centering
\includegraphics[width=1\textwidth]{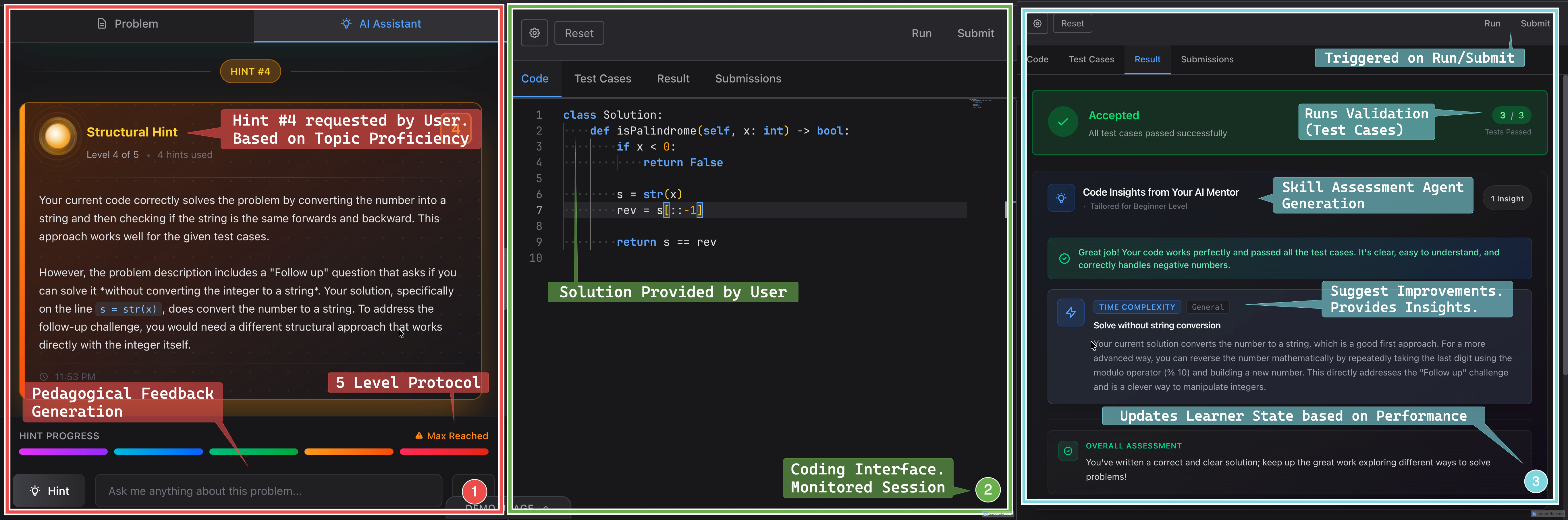}
\caption{Learner interface featuring the graduated hinting mechanism (Left), the Code Editor (Middle), and proficiency-aware code analysis (Right).}
\label{fig:interface}
\end{figure*}
\section*{System Demonstration}
\label{sec:demo}
The specificity of hints scales with the learner’s proficiency estimate $\hat{p}$. Beginners receive simple analogies and single-step cues, intermediate learners receive pattern-oriented guidance, and advanced learners receive concise nudges with edge-case emphasis. For the same recursion bug, a beginner might be told to “think of recursion like climbing down a ladder,” while an advanced learner might be prompted to “check whether the termination condition is reachable.”

\paragraph{Content Curator}
The Curator operationalizes the learner state into task selection using a dependency-aware 40/50/10 policy:
\begin{equation}
\scriptsize
\begin{split}
\text{selection} =\;& 0.4\,\times \text{due\_reviews} + 0.5\,\times \text{growth\_zone} + 0.1\,\times \text{challenge}.
\end{split}
\end{equation}

Growth-zone items correspond to mastery levels between $0.3$ and $0.7$, while challenge items target skills below $0.3$. The Curator enforces prerequisite dependencies, avoids repetition within a $k$-day window, and ensures topic diversity. For example, a learner showing intermediate recursion mastery but weak dynamic programming mastery may receive (i) a recursion review task, (ii) a medium-difficulty DP subproblem, and (iii) a lightweight DP challenge problem.

\paragraph{Progress Synthesizer}
The Progress Synthesizer governs spaced repetition using SM-2~\cite{wozniak1990optimization} and forgetting-curve theory~\cite{ebbinghaus1913memory}, augmented with contextual features~\cite{settles2016trainable, reddy2016unbounded}. Review intervals shrink when hints are heavily used, expand when solutions are fast and confident, and tighten when predicted recall drops near the due date. If the recall probability for a graph traversal concept falls below threshold, the agent prepones the review—even if the learner has not recently interacted.

\paragraph{Engagement Orchestrator}
Finally, the Engagement Orchestrator monitors motivational signals. It issues supportive prompts after broken streaks, encourages re-engagement after periods of inactivity, and suggests simpler variants when failure streaks accumulate. All interventions are rate-limited and phrased non-judgmentally. For instance, after multiple failed attempts on tree problems, the system may suggest: “Would you like to revisit the easier ‘binary tree basics’ exercise before trying again?”

Collectively, these agents form IntelliCode’s adaptation engine, grounding every hint, problem selection, and review decision in a unified, auditable learner state.

The IntelliCode platform is implemented using FastAPI for the backend, React for the frontend interface, and LangGraph for multi-agent orchestration. A persistent, graph-structured learner model is maintained in ArangoDB, enabling the system to track mastery, misconceptions, and review schedules across sessions.

The demo showcases the full adaptive tutoring loop. A learner begins at a curriculum roadmap and is assigned a data structures and algorithms (DSA) problem selected by the Content Curator using the 40/50/10 policy. If the learner struggles, the Pedagogical Feedback agent produces a proficiency-aligned hint—for example, a Level~2 conceptual cue such as “This problem requires identifying the recurrence pattern.” After incorporating the hint, the learner submits a correct solution. This submission triggers a sequence of coordinated agent behaviors. The Skill Assessment agent validates correctness and offers semantic feedback; the Learner Profiler updates the learner’s mastery estimate for recursion; and the Progress Synthesizer schedules a spaced-repetition review two days later, reflecting the hint usage and solve-time profile. The system interface allows the learner to view mastery trajectories, upcoming reviews, and past interactions, making the adaptation process transparent. This end-to-end interaction exemplifies how IntelliCode integrates real-time assessment, graduated hinting, curriculum adaptation, and spaced repetition into a coherent, state-driven teaching cycle.

\section{Evaluation Protocols}
\label{sec:evaluation}

We evaluate IntelliCode along offline, online, and fairness dimensions to demonstrate the reliability of its learner modeling, content adaptation, and multi-agent interactions.

\paragraph{Offline Metrics}
Our offline analysis focuses on validating the fidelity of the learner model. We measure \textbf{Mastery Calibration} by correlating predicted mastery $\hat{m}_t$ with subsequent correctness $y_{t+1}$ using Brier scores and Expected Calibration Error (ECE). The \textbf{Content Policy} is assessed by tracking topic coverage and diversity, with a target of at least $90\%$ coverage over a 30-day horizon to ensure balanced curriculum exposure. Finally, we examine \textbf{Scheduling Quality} by evaluating the accuracy of the recall predictor (AUROC $\geq 0.75$) and verifying adherence to SM-2 review due dates.

\paragraph{Online Metrics}
We also track performance in live interactions. \textbf{Learning Gains} are estimated through pre/post mastery changes on held-out assessments. \textbf{Engagement} is monitored through streak retention, inactivity gaps, and voluntary practice rates. System \textbf{Efficiency} is assessed via median end-to-end latency, with a target of under 500\,ms. \textbf{Safety} is evaluated through hint acceptance rates (aiming for $\geq 70\%$) and verification that the system never discloses full solutions.

\paragraph{Fairness Analysis and Ablations}
To ensure equitable support across learner profiles, we compare learning gains, hint levels, and pacing behaviors across proficiency deciles, targeting an interquartile range within $15\%$ of the median. We conduct ablation studies to isolate the contribution of key components, namely the Content Curator, Pedagogical Feedback agent, and context-aware SM-2 scheduler, quantifying their individual effects on calibration, mastery gains, and engagement.

\begin{figure}[h]
\centering
\includegraphics[width=1.0\linewidth]{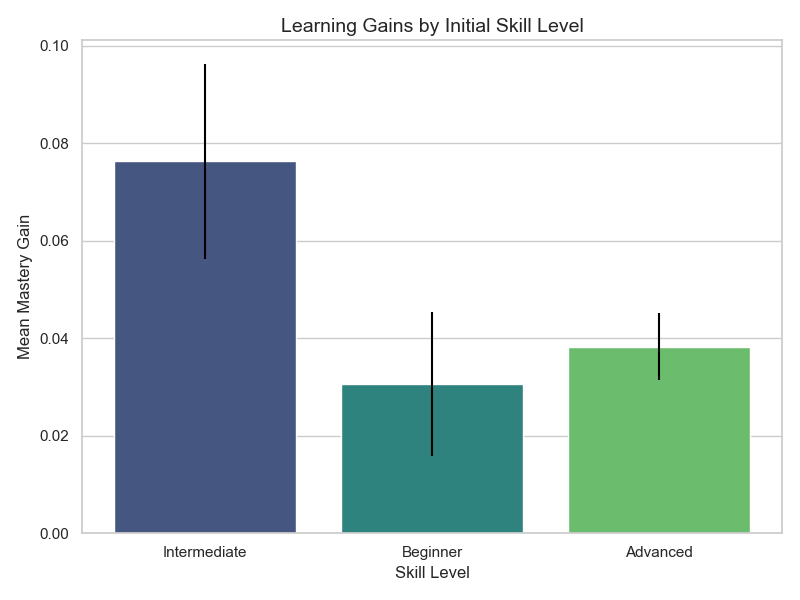}
\caption{Average mastery gains by initial skill level. Intermediate learners showed the highest sensitivity to the adaptive curriculum.}
\label{fig:learning_gains}
\end{figure}

\begin{figure}[h]
\centering
\includegraphics[width=0.7\linewidth]{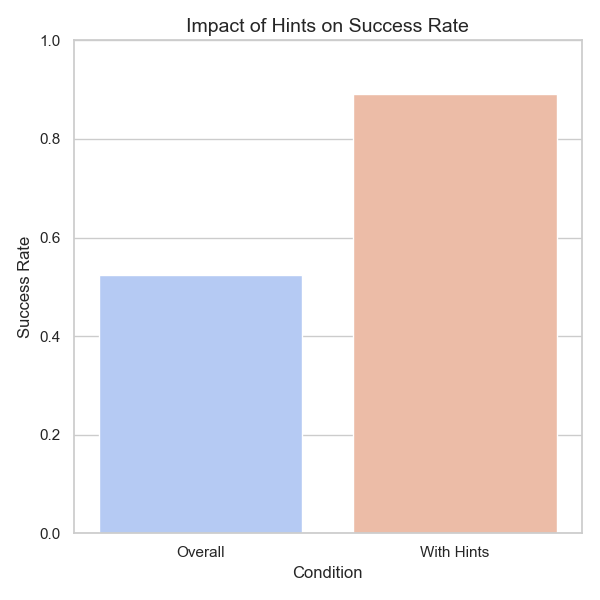}
\caption{Success rates with and without hint utilization.}
\label{fig:hint_effectiveness}
\end{figure}

\begin{figure}[h]
\centering
\includegraphics[width=1.0\linewidth]{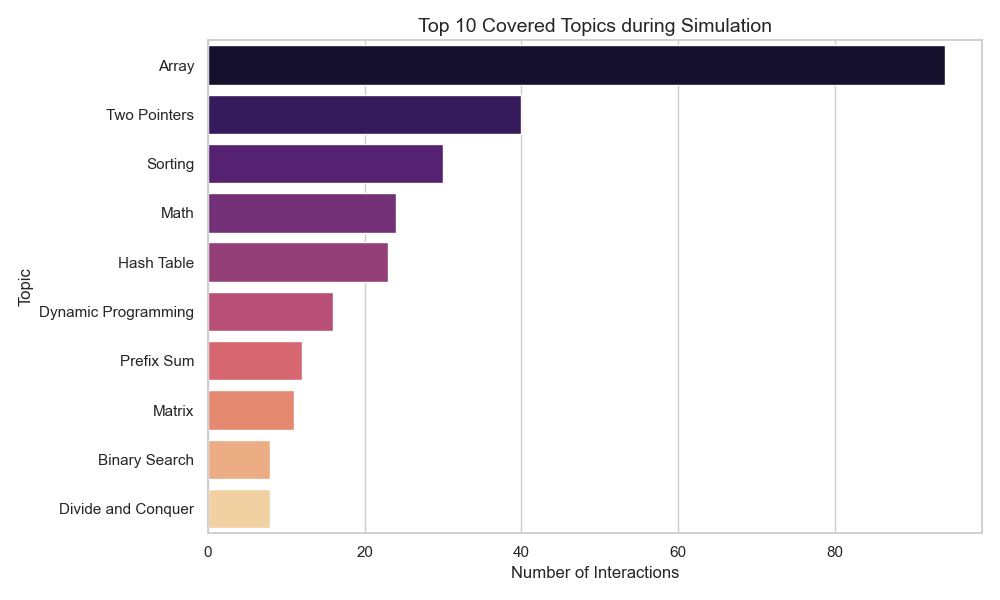}
\caption{Top 10 topics covered during the simulation.}
\label{fig:topic_dist}
\end{figure}

\subsection*{Validation with Simulated Learners}

To assess the responsiveness and stability of the multi-agent architecture, we conducted preliminary simulations using agent-based learner personas~\cite{wu2025embracingimperfectionsimulatingstudents}, drawing on methodologies from generative agent societies~\cite{park2023generativeagentsinteractivesimulacra}. While these simulations cannot substitute for human studies in evaluating educational efficacy, they provide an initial validation of the system’s ability to respond coherently to diverse cognitive profiles and interaction patterns.

\subsection{Responsiveness and Hint Effectiveness}

Across ten simulated learning trajectories, IntelliCode dynamically adapted task difficulty and hinting behavior, producing a mean mastery gain of 5.04\% (Figure~\ref{fig:learning_gains}). The graduated hinting mechanism also behaved robustly: tasks in which simulated learners requested hints exhibited a success rate of 89.1\%, compared with an overall baseline of 52.4\% (Figure~\ref{fig:hint_effectiveness}). These results confirm that the Pedagogical Feedback agent intervenes effectively offering conceptual guidance without disclosing solutions while maintaining architectural reliability.

\subsection{Content Coverage}

The Content Curator maintained strong diversity across topics (Figure~\ref{fig:topic_dist}), demonstrating the ability of the 40/50/10 policy to avoid topic starvation while respecting prerequisite relationships. Even in extended sessions, the system preserved balanced coverage across skill areas, validating the orchestrator’s ability to coordinate long-term learning arcs and manage curriculum progression.

\section{Conclusion}\label{sec:conclusion}

\label{sec:conclusion}

In this paper, we presented IntelliCode, a principled multi-agent LLM framework for adaptive education built around a persistent, auditable learner state. By integrating formal mastery update rules, proficiency-aware graduated hinting, dependency- and fairness-aware curriculum adaptation, and safety-aligned prompting, the system offers transparent and consistent multi-turn tutoring capabilities. The simulations demonstrate stable architectural behavior, effective hinting interventions, and robust content coverage, highlighting the technical viability of the approach. 
We envision IntelliCode as a foundation for next-generation educational systems that blend modern LLM reasoning with established principles from learner modeling, instructional design, and cognitive science.



\section*{Limitations}
\label{sec:limitations}

While IntelliCode demonstrates promising architectural and pedagogical capabilities, there are a few limitations. First, the mastery estimates rely on BKT/DKT-inspired proxies that require sufficient interaction scale and careful calibration; cold-start learners, in particular, necessitate conservative priors and may experience reduced personalization in early sessions. Second, LLM-driven components introduce variability due to model drift, occasional refusals, and cost constraints. Our guardrails and validation schemas mitigate these issues but cannot fully eliminate them. Finally, rigorous fairness evaluation requires diverse and representative datasets, and remains vulnerable to selection bias, behavior-signal noise, LLM drift, data leakage, and survivorship bias. These considerations underscore the need for large-scale, longitudinal human studies in future work.

\section*{Ethical Considerations}
\label{sec:ethics}

The deployment of LLM-based agents in educational settings necessitates careful attention to accuracy, dependency, and privacy. While IntelliCode integrates verifiers like the \textit{Skill Assessment} agent to validate code logic, generative components such as the \textit{Pedagogical Feedback} agent remain susceptible to hallucinations or plausible but incorrect explanations. Consequently, the system is designed to function as a supplemental tutor rather than a replacement for formal instruction, and we recommend its use under the guidance of human educators who can monitor for potential deviations.

To mitigate the risk of learner over-dependence on AI assistance, we implemented strict graduated hinting protocols. However, we acknowledge that prolonged reliance on automated scaffolding may impact unassisted problem-solving capabilities. Our design prioritizes metacognitive prompting over direct solution disclosure to foster genuine skill acquisition.

Regarding data privacy, all learner interactions are processed with strict minimization principles. Personally Identifiable Information (PII) is redacted prior to agent ingestion, and the \textit{Learner Profiler} operates on anonymized mastery integers rather than raw user profiles. Finally, while our curriculum policy incorporates fairness constraints to ensure equitable topic coverage, the underlying datasets used for cold-start calibration may inherently reflect historical biases, requiring ongoing monitoring of learning outcomes across diverse demographic groups.

\vspace{1em}
\noindent\textit{\textbf{Submission Note:} This paper has been submitted to EACL 2026 System Demonstrations Track for review.}
\vspace{0.5em}

\bibliography{custom}

\appendix

\section{Resources and Availability}
\label{app:resources}

To support reproducibility and further research, we provide open access to our platform components and simulation data under the \textbf{MIT License}:

\begin{itemize}
    \item \textbf{Live Demo:} \url{https://intellicode.redomic.in}
    \item \textbf{Frontend:} \url{https://github.com/Redomic/Intellicode-frontend}
    \item \textbf{Backend:} \url{https://github.com/Redomic/Intellicode-backend}
    \item \textbf{Simulation Framework:} \url{https://github.com/Redomic/intellicode_student_sim}
\end{itemize}

\section{Mathematical Formulation Details}\label{app:formulation}

We formulate adaptive personalized education as a Partially Observable Markov Decision Process (POMDP):
\begin{equation}
\text{POMDP} = (\mathcal{S}, \mathcal{A}, \mathcal{O}, T, R, \gamma, b_0)
\end{equation}

\subsection{State Space}

The learner state at time $t$ is:
\begin{equation}
\mathbf{S}_t = \{m_t, r_t, e_t, p_t, M_t, v_t\}
\end{equation}
where:
\begin{itemize}
\item $m_t$: mastery vector, $m_{t,i} \in [0,1]$ for topic $i \in T$
\item $r_t$: review schedule, items with $(q_{\text{id}}, \text{topics}, d_{\text{due}}, \text{interval}, \text{EF}, n_{\text{reviews}})$
\item $e_t$: engagement state, streak, last-seen timestamp, recent activity windows
\item $p_t$: preferences, skill level, modality, time budget, opt-outs
\item $M_t$: long-term memory, structured text sections on trends, misconceptions, insights
\item $v_t$: version, timestamp for auditing
\end{itemize}

We also maintain uncertainty $u_{t,i}$ per topic, encoded as Beta parameters $(\alpha_{t,i}, \beta_{t,i})$.

\subsection{Observation Space}

Observations are partial, noisy signals of state:
\begin{equation}
\begin{split}
\mathcal{O}_t \in \{\,\text{submission},\; 
\text{hint\_request},\\
\text{session\_start},\;
\text{due\_review}\,\}
\end{split}
\end{equation}

For each submission, we observe:
\begin{equation}
o_t = (q_{\text{id}}, y, \tau, h_{\text{cnt}}, \text{errors}, t_{\text{solve}})
\end{equation}
where $y \in \{0,1\}$ (pass/fail), $\tau$ (timestamp), $h_{\text{cnt}}$ (hints used), errors (semantic signals), $t_{\text{solve}}$ (time on task).

\subsection{Action Space}

The orchestrator (via agents) selects actions:
\begin{equation}
\begin{split}
\mathcal{A}_t \in \{&
\text{recommend\_item},\;
\text{hint}(l),\;\\
&\text{adjust\_schedule},
\text{intervene},\;
\text{feedback}(d)
\}
\end{split}
\end{equation}

where $l \in \{1, 2, 3, 4, 5\}$ is hint level, $d$ is feedback detail level.

\subsection{Reward Function}

The reward proxies learning progress while penalizing inefficiency:
\begin{equation}
\begin{split}
R_t =\;& \underbrace{w_m \Delta m_t}_{\text{mastery gain}}
+ \underbrace{w_r \mathbb{1}[\text{review\_success}]}_{\text{retention}} \\
&- \underbrace{w_h h_{\text{cnt}}}_{\text{hint penalty}}
- \underbrace{w_t \max(0, t_{\text{solve}} - \mu_t)}_{\text{time penalty}}
\end{split}
\end{equation}

with regularizers for fairness (topic coverage) and engagement.

\section{Learner State Updates}\label{app:learner_state}

\subsection{State Initialization}

From historical submissions, we compute initial mastery using a recency-weighted exponential moving average:
\begin{equation}
\begin{split}
m_{t,i}^{(0)} =\;& 0.6\,\text{success\_rate}_i \\
&+ 0.4\,\text{recent\_success\_rate}_i \\
&+ \mathcal{N}(0, \sigma_0^2)
\end{split}
\end{equation}
We initialize Beta parameters as $(\alpha_0, \beta_0) = (1, 1)$ (uninformative prior), and review queue empty with ease factor $\text{EF}_0 = 2.5$.

\subsection{Mastery Update Rule}

Upon outcome $y \in \{0,1\}$ on a question tagged with topics $Q$:
\begin{equation}
m_{t,i} \gets 
\begin{cases}
\min(1,\; m_{t,i} + \alpha w_d w_r (1 - m_{t,i})) \\[-3pt]
\quad \scriptstyle (y=1),\\[4pt]
\max(0,\; m_{t,i} - \beta w_d^{-1} w_r m_{t,i}) \\[-3pt]
\quad \scriptstyle (y=0)
\end{cases}
\end{equation}

where:
\begin{itemize}
\item $w_d \in \{0.8, 1.0, 1.2\}$ maps difficulty $\in \{\text{Easy}, \text{Medium}, \text{Hard}\}$ (inverted for failure penalties)
\item $w_r = \exp(-\Delta t / \tau_{\text{upd}})$ decays with recency
\item Hint/time penalties: $m_{t,i} \gets m_{t,i} - \eta_h h_{\text{used}} - \eta_t \max(0, t_{\text{solve}} - \mu_i)$
\item Momentum smoothing: $m_{t,i} \gets (1-\lambda) m_{t,i} + \lambda m_{t,i}^{\text{new}}$ reduces jitter
\end{itemize}

\subsection{Proficiency Composite}

Overall proficiency is a weighted composite:
\begin{equation}
\hat{p} = \sum_k w_k s_k
\end{equation}
where $s_k$ includes: topic mastery average (0.40), expertise rank (0.25), self-reported skill (0.20), recent success rate (0.10), streak normalization (0.05).

\subsection{Spaced Repetition Updates (SM-2)}

For a review item with ease factor $\text{EF}$, we derive quality score $q \in \{0,1,2,3,4,5\}$ from:
\begin{equation}
q = \begin{cases}
5 & \text{fast, no hints} \\
4 & \text{solved, minor delay} \\
3 & \text{solved with hints} \\
\leq 2 & \text{failed/forgot}
\end{cases}
\end{equation}
Update: 
\begin{equation}
\text{EF}' = \max(1.3, \text{EF} - 0.8 + 0.28q - 0.02q^2)
\end{equation}
Intervals: $I_1 = 1$, $I_2 = 6$, $I_n = \text{round}(I_{n-1} \cdot \text{EF}')$ days.
Predicted recall: 
\begin{equation}
R(\Delta t) = \exp(-\Delta t / \tau)
\end{equation}
where $\tau \propto \text{EF}'$.

\section{Reproducibility Details}
\label{sec:reproducibility}

\begin{enumerate}
\item Seeded random splits; fixed topic-graph snapshot.
\item Frozen prompt versions and role texts.
\item Logged hyperparameters: $\alpha, \beta, w_d, \tau_{\text{upd}}, \lambda$, proficiency weights.
\item Validation schemas for agent outputs (JSON specs).
\item Behavior signal preprocessing with masking rules.
\item Ablation code and offline evaluation scripts.
\end{enumerate}
\end{document}